 \documentclass[pmlr,twocolumn,10pt]{jmlr} 

\usepackage{booktabs}
\usepackage{makecell}
\usepackage{bm} 
\usepackage[load-configurations=version-1]{siunitx} 

\usepackage{colortbl}

\usepackage{todonotes}
\usepackage{float}
\usepackage[format=plain]{caption}
\usepackage{multicol}

\theorembodyfont{\upshape}
\theoremheaderfont{\scshape}
\theorempostheader{:}
\theoremsep{\newline}

\jmlrvolume{LEAVE UNSET}
\jmlryear{2021}
\jmlrsubmitted{LEAVE UNSET}
\jmlrpublished{LEAVE UNSET}
\jmlrworkshop{Machine Learning for Health (ML4H) 2021} 

\title[Monte Carlo dropout increases model repeatability]{Monte Carlo dropout increases model repeatability}

\author{%
\Name{Andreanne Lemay}\nametag{$^{1, 2, 3}$} \Email{andreanne.lemay@polymtl.ca}\\
\addr $^{1}$ Martinos Center for Biomedical Imaging, Department of Radiology, Massachusetts General Hospital, Boston, USA \\
\addr $^{2}$ NeuroPoly Lab, Institute of Biomedical Engineering, Polytechnique Montreal, Canada \\
\addr $^{3}$ Mila, Quebec AI Institute, Canada \\
\vspace{-3mm}
\AND
\Name{Katharina Hoebel}\nametag{$^{1,4}$} \Email{khoebel@mit.edu}\\
\addr $^{4}$ Massachusetts Institute of Technology, Cambridge, USA \\
\vspace{-3mm}
\AND
\Name{Christopher P. Bridge}\nametag{$^{1, 5}$} \Email{cbridge@partners.org}\\
\addr $^{5}$ MGH and BWH Center for Clinical Data Science, Boston, USA \\
\vspace{-3mm}
\AND
\Name{Didem Egemen}\nametag{$^{6}$} \Email{didem.egemen@nih.gov}\\
\addr $^{6}$ National Cancer Institute, Clinical Genetics Branch, Division of Cancer Epidemiology {\&} Genetics, Rockville, USA \\
\vspace{-3mm}
\AND
\Name{Ana Cecilia Rodriguez}\nametag{$^{6}$} \Email{rodriguezac2@gmail.com}\\
\vspace{-3mm}
\AND
\Name{Mark Schiffman}\nametag{$^{6}$} \Email{schiffmm@exchange.nih.gov}\\
\vspace{-3mm}
\AND
\Name{John Peter Campbell}\nametag{$^{7}$} \Email{campbelp@ohsu.edu}\\
\addr $^{7}$ Oregon Health and Science University, Portland, USA \\
\vspace{-3mm}
\AND
\Name{Jayashree Kalpathy-Cramer}\nametag{$^{1}$}  \Email{kalpathy@nmr.mgh.harvard.edu}\\
\vspace{-12mm}
}

\begin{document}

\maketitle

\begin{abstract}
The integration of artificial intelligence into clinical workflows requires reliable and robust models. Among the main features of robustness is repeatability. Much attention is given to classification performance without assessing the model repeatability, leading to the development of models that turn out to be unusable in practice. In this work, we evaluate the repeatability of four model types on images from the same patient that were acquired during the same visit. We study the performance of binary, multi-class, ordinal, and regression models on three medical image analysis tasks: cervical cancer screening, breast density estimation, and retinopathy of prematurity classification. Moreover, we assess the impact of sampling Monte Carlo dropout predictions at test time on classification performance and repeatability. Leveraging Monte Carlo predictions significantly increased repeatability for all tasks on the binary, multi-class, and ordinal models leading to an average reduction of the 95\% limits of agreement by 17\% points.
\end{abstract}


\section{Introduction}
\begin{figure}[htp]
  \begin{center}
    \subfigure[\small{Model prediction: 0.01 (Normal)}]{\label{fig:left_cervix}\includegraphics[width=0.39\linewidth]{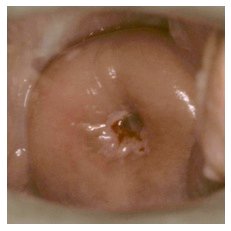}}
    \subfigure[\small{Model prediction: 0.98  (Pre-cancer)}]{\label{fig:right_cervix}\includegraphics[width=0.39\linewidth]{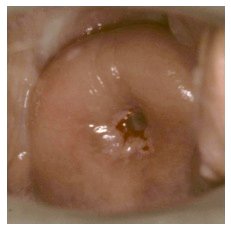}} \\
  \end{center}
  \vspace{-3mm}
  \caption{\small{\textbf{Illustration of repeatability issues from DL models on different images of a cervix with precancerous lesions from the same patient taken the same day.} (a) the binary model (non MC) predicts a normal cervix (severity score: 0.01). (b) the binary model predicts pre-cancer (severity score: 0.98).}}

  \label{fig:problem}

\end{figure}

\vspace{-2mm}
While the majority of the deep learning (DL) work for medical image classification focuses on accuracy and classification performance, little is known about the repeatability of these models. It is of utmost importance and key to model reliability, that models provide consistent predictions in clinical settings. Empirically, minor changes in an image can lead to vastly different predictions by DL models. In clinical practice, this repeatability issue could lead to dangerous medical errors. Figure \ref{fig:problem} illustrates this issue. Automatic visual assessment of cervix images by artificial intelligence models has the potential to support cervical cancer screening in low-resource settings. Two cervical cancer screening images from the same precancerous cervix acquired during the same visit lead to completely different predictions. A binary DL model (without dropout layers) trained to distinguish between a normal cervix and one with precancerous lesions (0: Normal, 1: Pre-cancer) predicted a normal cervix on one image and classified the second one as precancerous. This difference is represented by prediction results at each extreme of the spectrum, i.e., 0.01 and 0.98, suggesting high model certainty for both outputs. Repeatability is important as it reflects how reliably the model can achieve a certain classification performance.

\begin{table*}[ht]
\begin{center}
\vspace{-11mm}
\floatconts{tab:dataset}
  {\caption{\small{\textbf{Datasets description}. Mean image per patient is the mean number of images available for the same anatomical structure at the same timepoint. img: image; No.: Normal; Scat.: scattered; Hetero: Heterogeneous.}} \vspace{-5mm}}
   
  {\begin{tabular}{lcccc}
  \toprule
  & \# images & Label distribution & \thead{Mean \\ img/patient} & Train/Val/Test split \\
  \midrule
  Cervical & 3509 & No./Gray/Precancer \small{(33\%/33\%/34\%)} & 2 & 65\%/10\%/25\% \\
  DMIST & 108,230 & Fatty/Scat./Hetero./Dense \small{(12\%/44\%/38\%/6\%)}  & 5 & 65\%/10\%/25\% \\
  ROP & 5511 & No./Pre-Plus/Plus \small{(82\%/15\%/3\%)}  & 3 & 78\%/13\%/8\% \\
  \bottomrule
  \end{tabular}}
  \end{center}
  \vspace{-7mm}
\end{table*}
\vspace{-5mm}
\subsection{Study outline}
In the competition to achieve higher and higher classification performance important aspects like test-retest variability remain overlooked and not all DL models are equal with respect to their repeatability. To the best of our knowledge, this is the first study to propose Monte Carlo (MC) dropout \citep{hinton2012improving, srivastava2014dropout} as a method to improve repeatability and systematically assess it on different tasks, model types, and network architectures. In this work, we evaluate the repeatability of four types of DL classification and regression models, binary, multi-class, ordinal, and regression, each with and without MC dropout. We test the repeatability of these models' predictions on three different medical image analysis tasks: cervical cancer screening, breast density estimation, and retinopathy of prematurity (ROP) disease severity classification. Based on our results, we present recommendations for model selection to improve repeatability.

\vspace{-2mm}
\section{Methods}
\vspace{-1mm}

\begin{figure*}[ht]

\vspace{-10mm}
\begin{center}
  {\includegraphics[width=0.85\linewidth]{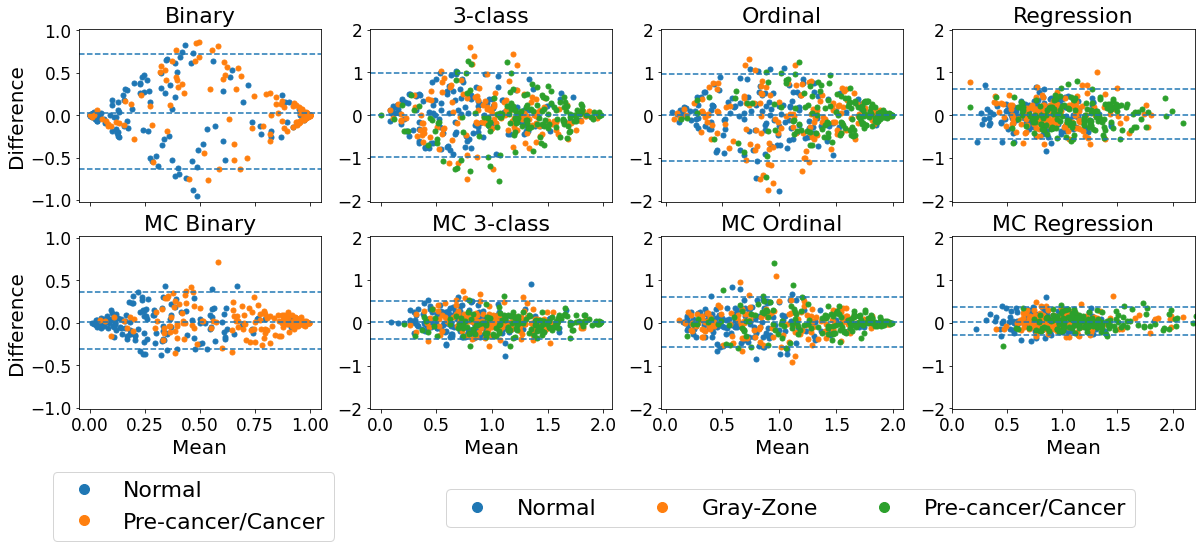}}
  \floatconts{fig:bland_altman}
  {\vspace{-5mm} \caption{\small{\textbf{Bland-Altman plots on images from the same patient and visit for cervical classification models.} The y-axis of each graph represents the difference in model prediction for images of the same patient, while the x-axis refers to the mean of the predicted scores. The 95\% limits of agreement are presented with dashed lines.}}}
  
 \end{center}
\vspace{-3mm}
\end{figure*}

\begin{table*}[ht]
\begin{center}
\floatconts{tab:metric-table}
  {\caption{\small{\textbf{Model performance overview (MEAN $\pm$ 95\% CI)}. Values in bold indicate statistical difference between MC and non-MC models. LoA: Limits of agreement; Acc.: Accuracy; CI: Confidence interval. }} \vspace{-5mm}}
   
  {\begin{tabular}{lcc|cc|cc}
  \toprule
  & \multicolumn{2}{c}{\textbf{Cervical}} & \multicolumn{2}{c}{\textbf{Breast density}} & \multicolumn{2}{c}{\textbf{ROP}}\\
  \midrule
  \textbf{Model} & \textbf{95\% LoA} $\downarrow$ & \textbf{Acc.} $\uparrow$  & \textbf{95\% LoA} $\downarrow$ & \textbf{Acc.} $\uparrow$  & \textbf{95\% LoA} $\downarrow$ & \textbf{Acc.} $\uparrow$  \\
  \midrule
  \small{Binary} & \small{$0.68 \pm 0.07$} & \small{$0.73 \pm 0.03$} & \small{$0.58 \pm 0.01$} & \small{$0.84 \pm 0.00$}  & \small{$0.88 \pm 0.04$} & \small{$0.81 \pm 0.02$} \\
  \small{MC Binary} & \small{\bm{$0.33 \pm 0.04$}} & \small{\bm{$0.75 \pm 0.03$}} & \small{\bm{$0.48 \pm 0.01$}} & \small{\bm{$0.85 \pm 0.00$}} & \small{\bm{$0.55 \pm 0.05$}} & \small{\bm{$0.85 \pm 0.02$}}\\
 \arrayrulecolor{gray}\hline
  \small{Multi class} & \small{$0.50 \pm 0.06$} &  \small{$0.47 \pm 0.03$} & \small{$0.33 \pm 0.00$} & \small{$0.69 \pm 0.01$} & \small{$0.48 \pm 0.03$} & \small{$0.85 \pm 0.02$}\\
  \small{MC multi-class} & \small{\bm{$0.22 \pm 0.03$}} & \small{\bm{$0.52 \pm 0.03$}} & \small{\bm{$0.30 \pm 0.00$}} & \small{\bm{$0.71 \pm 0.01$}} & \small{\bm{$0.39 \pm 0.03$}} & \small{\bm{$0.85 \pm 0.02$}} \\
  \arrayrulecolor{gray}\hline
  \small{Ordinal} & \small{$0.51 \pm 0.07$} & \small{$0.47 \pm 0.03$} & \small{$0.33 \pm 0.00$} & \small{$0.68 \pm 0.01$} & \small{$0.40 \pm 0.04$} & \small{$0.82 \pm 0.02$} \\
  \small{MC ordinal} & \small{\bm{$0.29 \pm 0.03$}} & \small{\bm{$0.49 \pm 0.03$}} & \small{\bm{$0.29 \pm 0.01$}} & \small{\bm{$0.69 \pm 0.01$}} & \small{\bm{$0.34 \pm 0.03$}} & \small{\bm{$0.83 \pm 0.02$}}\\
  \arrayrulecolor{gray}\hline
  \small{Regression} & \small{$0.29 \pm 0.03$} & \small{\bm{$0.44 \pm 0.03$}} & \small{$0.21 \pm 0.01$} & \small{\bm{$0.70 \pm 0.01$}} & \small{\bm{$0.33 \pm 0.03$}} & \small{\bm{$0.86 \pm 0.02$}}  \\
  \small{MC Regression} & \small{\bm{$0.16 \pm 0.02$}} & \small{$0.43 \pm 0.03$} & \small{$0.21 \pm 0.01$} & \small{$0.67 \pm 0.01$} & \small{$0.33 \pm 0.01$} & \small{$0.79 \pm 0.03$} \\
    \arrayrulecolor{black}
  \bottomrule
  \end{tabular}}
   \end{center}
  \vspace{-5mm}
\end{table*}

\subsection{Datasets}

Three datasets were used to assess the repeatability of the different models: cervical photographs for certical cancer screening, mammographies for breast density prediction, and retinal photographs for ROP severity diagnosis (summarized in Table \ref{tab:dataset}). All datasets were split on a patient level to prevent leakage of information. Repeatability was assessed as the maximum difference between model outputs for all available images for each patient in the test dataset. 

\paragraph{Cervical} Cervical cancer is the fourth most common cancer in women worldwide and the leading cause of cancer-related deaths in western, eastern, middle, and southern Africa \citep{Arbyn2020EstimatesAnalysis}. In addition to HPV testing, the visual assessment of the cervix using photographs can help to detect precancerous lesions in low-resource settings \citep{Catarino2015CervicalChoices, Xue2020ACamera, Hu2019AnScreening}. Each image, originating from one of two studies \citep{schiffman2003findings, bratti2004description}, was classified using cytological and histological data from the patient as one of the following three categories: Normal,  Gray zone, i.e., the presence of precancerous lesions was equivocal, Pre-cancer/cancer. For the binary models, we utilized only images that were classified as either normal or pre-cancer/cancer.

\paragraph{Breast density} The risk to develop breast cancer rises with increasing breast density \citep{Boyd1995QuantitativeStudy}. We used images from the Digital Mammographic Imaging Sceening Trial (DMIST) acquired at 33 institutions \citep{Pisano2005DiagnosticScreening}. Breast density labels were generated according to the BI-RADS criteria \citep{Liberman2002BreastBI-RADS}: fatty, scattered fibroglandular density, heterogeneously dense, and extremely dense. The breast density models were trained using all available views for each patient in the training dataset using either four labels or a simplified binary labelling system of fatty and scattered as one class, and dense and heterogeneous as the other class. 

\paragraph{Retinopathy of Prematurity}
ROP is the leading cause of preventable childhood blindness worldwide \citep{BlindnessVISION2020}. Images from the dataset described in \citet{Brown2018AutomatedNetworks} were classified as normal, pre-plus, or plus disease following previously published methods \citep{Ryan2014DevelopmentOphthalmology}. The final label was based on the independent image-based diagnosis by 3 expert graders. The binary models were trained using normal as one class and pre-plus and plus as the other class. Following \citet{Brown2018AutomatedNetworks}'s work, we trained ROP classification models using pre-segmented vessel maps as input. ROP classification models were trained using only the posterior field of view and the repeatability was assessed on the posterior, temporal, and nasal views which represents a slight domain shift (i.e., testing on unseen views). 

\paragraph{Classification model training}
For each dataset, we trained binary, multi-class and ordinal \citep{cao2019rank} classification models, as well as regression models each with and without dropout, resulting in a total of 8 models per dataset. 
Models with dropout were trained using spatial dropout with a dropout rate of 0.1 for cervical images and DMIST, and 0.2 for ROP. For the DenseNet121 architecture, the dropout was applied after every dense layer while for the ResNets the dropout layer was applied after each residual block. At test time, the dropout was enabled to generate $N=50$ slightly different predictions and the final prediction was obtained by averaging over all the MC samples \citep{gal2016dropout}.
We used the following ImageNet pretrained models for each dataset based on which performed the best for the conventional multi-class classification model: DenseNet121 (cervix), ResNet50 (DMIST), and ResNet18 (ROP). Models were trained using binary cross-entropy, cross-entropy, CORAL \citep{cao2019rank}, and mean squared error (MSE) losses for binary, multi-class, ordinal, and regression models respectively. The code was implemented using the MONAI framework \citep{the_monai_consortium_2020_4323059} based on the PyTorch library \citep{paszke2017automatic}.

\subsection{Evaluation}
\paragraph{Severity scores} 

For direct comparison of a model's predictions, we summarized each model's outputs as a continuous severity score. The methodology and analysis were chosen based on the consideration that the underlying variable of interest, i.e.  disease severity, of these medical tasks is better represented by a spectrum rather than clear distinct categories. For the binary and regression models, the output of the models was directly used without further modifications (labels from 0 to max class - 1). For the multi-class model, we utilized the ordinality of all three classification problems and defined the continuous severity score as a weighted average using softmax probability of each class as described in Equation \ref{eq:multiclass}. For cervical and ROP classification (3 classes), the values lie in the range 0 to 2 while the values extend from 0 to 3 for breast density (4 classes).
\vspace{-3mm}
 \begin{equation} \label{eq:multiclass}
    score = \sum_{i=1}^{k} p_{i} \times i - 1
 \end{equation}

\noindent with $k$ being the number of classes and $p_{i}$ the softmax probability of class $i$.
For the ordinal model, the classification problem of $k$ ranks (i.e., classes) is modified into $k - 1$ binary classification problems \citep{li2007ordinal} leading to one output unit less than for the classification form. For instance, for a 3-class problem, the ground truth would be encoded as follows: class 1 $\rightarrow$ [0, 0]; class 2 $\rightarrow$ [1, 0]; class 3 $\rightarrow$ [1, 1]. The continuous prediction score for ordinal models is obtained by summing the output neurons leading to values ranging from 0 to 2 and 0 to 3 for 3-class and 4-class problems, respectively.

\paragraph{Metrics} Repeatability was evaluated using the 95\% limits of agreement (LoA) from the Bland-Altman plots. Since normality was not reached for the differences for the LoA, non-parametric LoA were calculated using empirical percentiles \citep{bland1999measuring}. The LoA was presented as a fraction of the possible value range. The classification accuracy was also reported. For the regression models, thresholds to binarize predictions for accuracy calculation were computed by splitting the range of predictions equally (e.g., 3-class problem with predictions ranging from 0 to 2: $s \leq 0.67 \rightarrow$ class 1; $0.67 < s \leq 1.33 \rightarrow$ class 2; $s \geq 1.33 \rightarrow$ class 3).

\paragraph{Statistics} Statistical difference between models was determined using a two-sided $t$-test and metric bootstrapping (500 iterations). Models with a p-value smaller than 0.05 were considered significantly different. The normality of the distribution was verified using the Shapiro-Wilk test ($\alpha=0.05$).

\vspace{-3mm}
\section{Results}
\vspace{-2mm}
Bland-Altman plots of all model types for cervical classification are summarized in Figure \ref{fig:bland_altman}.  Ideally, all cases would lie near a horizontal line crossing the y-axis at 0 which means the difference between test-retest score is low. For every task, the MC models showed better test-retest reliability than their conventional counterparts with the exception of the regression models. This is illustrated by the narrower 95\% LoA and the highest concentration of differences near 0 on the y-axis. The range of predicted values remained similar for MC models, indicating that the effect of the MC model is not simply regressing scores towards the mean. 

 95\% LoA (i.e., repeatability metric) and accuracy for each model can be found in Table \ref{tab:metric-table}. As observed in Figure \ref{fig:bland_altman}, repeatability was statistically improved when using MC models for binary, multi-class and ordinal models for all tasks. On average, across all tasks and classification models (i.e., excluding regression) the 95\% limit of agreement improved by 17\% points. Increased repeatability translates to smaller disagreement rates during test-retest which is desired for classification tasks. Accuracy increased for all classification MC models. No consistent improvement in classification or repeatability performances was shown when adding MC iterations to regression models over the different medical tasks. Regression models generally showed better repeatability compared with the classification models (i.e., 3-class and ordinal), which reduces the potential to improve repeatability, as regression without MC reaches good repeatability performances compared to the classification models.

\vspace{-2mm}
\section{Conclusion}
\vspace{-2mm}
We evaluated the repeatability of four model types on three medical tasks using distinct model architectures (ResNet18, ResNet50, DenseNet121). We demonstrated that MC sampling during test time leads to more reliable classification models providing more stable and repeatable predictions on different images from the same patient. Only regression models did not show a constant improvement when leveraging MC sampling. MC sampling is flexible as it is applicable to any model type and architecture while being easily implementable. The code used to train and generate results can be found at \url{https://github.com/andreanne-lemay/gray_zone_assessment}.

\newpage
\bibliography{jmlr-sample,kathi_mendeley}

\begin{thebibliography}{20}
\providecommand{\natexlab}[1]{#1}
\providecommand{\url}[1]{\texttt{#1}}
\expandafter\ifx\csname urlstyle\endcsname\relax
  \providecommand{\doi}[1]{doi: #1}\else
  \providecommand{\doi}{doi: \begingroup \urlstyle{rm}\Url}\fi

\bibitem[Arbyn et~al.(2020)Arbyn, Weiderpass, Bruni, de~Sanjos{\'{e}}, Saraiya,
  Ferlay, and Bray]{Arbyn2020EstimatesAnalysis}
Marc Arbyn, Elisabete Weiderpass, Laia Bruni, Silvia de~Sanjos{\'{e}}, Mona
  Saraiya, Jacques Ferlay, and Freddie Bray.
\newblock {Estimates of incidence and mortality of cervical cancer in 2018: a
  worldwide analysis}.
\newblock \emph{The Lancet Global Health}, 8\penalty0 (2):\penalty0 e191--e203,
  2020.
\newblock ISSN 2214109X.
\newblock \doi{10.1016/S2214-109X(19)30482-6}.

\bibitem[Bland and Altman(1999)]{bland1999measuring}
J~Martin Bland and Douglas~G Altman.
\newblock Measuring agreement in method comparison studies.
\newblock \emph{Statistical methods in medical research}, 8\penalty0
  (2):\penalty0 135--160, 1999.

\bibitem[Boyd et~al.(1995)Boyd, Byng, Jong, Fishell, Little, Miller, Lockwood,
  Tritchler, and Yaffe]{Boyd1995QuantitativeStudy}
N.~F. Boyd, J.~W. Byng, R.~A. Jong, E.~K. Fishell, L.~E. Little, A.~B. Miller,
  G.~A. Lockwood, D.~L. Tritchler, and M.~J. Yaffe.
\newblock {Quantitative classification of mammographic densities and breast
  cancer risk: Results from the canadian national breast screening study}.
\newblock \emph{Journal of the National Cancer Institute}, 87\penalty0
  (9):\penalty0 670--675, 1995.
\newblock ISSN 00278874.
\newblock \doi{10.1093/jnci/87.9.670}.

\bibitem[Bratti et~al.(2004)Bratti, Rodr{\'\i}guez, Schiffman, Hildesheim,
  Morales, Alfaro, Guill{\'e}n, Hutchinson, Sherman, Eklund,
  et~al.]{bratti2004description}
M~Concepci{\'o}n Bratti, Ana~C Rodr{\'\i}guez, Mark Schiffman, Allan
  Hildesheim, Jorge Morales, Mario Alfaro, Diego Guill{\'e}n, Martha
  Hutchinson, Mark~E Sherman, Claire Eklund, et~al.
\newblock Description of a seven-year prospective study of human papillomavirus
  infection and cervical neoplasia among 10 000 women in guanacaste, costa
  rica.
\newblock \emph{Revista Panamericana de Salud P{\'u}blica}, 15:\penalty0
  75--89, 2004.

\bibitem[Brown et~al.(2018)Brown, Campbell, Beers, Chang, Ostmo, Chan, Dy,
  Erdogmus, Ioannidis, Kalpathy-Cramer, and Chiang]{Brown2018AutomatedNetworks}
James~M. Brown, J.~Peter Campbell, Andrew Beers, Ken Chang, Susan Ostmo,
  R.~V.Paul Chan, Jennifer Dy, Deniz Erdogmus, Stratis Ioannidis, Jayashree
  Kalpathy-Cramer, and Michael~F. Chiang.
\newblock {Automated diagnosis of plus disease in retinopathy of prematurity
  using deep convolutional neural networks}.
\newblock \emph{JAMA Ophthalmology}, 136\penalty0 (7):\penalty0 803--810, 2018.
\newblock ISSN 21686165.
\newblock \doi{10.1001/jamaophthalmol.2018.1934}.

\bibitem[Cao et~al.(2019)Cao, Mirjalili, and Raschka]{cao2019rank}
Wenzhi Cao, Vahid Mirjalili, and Sebastian Raschka.
\newblock Rank-consistent ordinal regression for neural networks.
\newblock \emph{arXiv preprint arXiv:1901.07884}, 1\penalty0 (6):\penalty0 13,
  2019.

\bibitem[Catarino et~al.(2015)Catarino, Petignat, Dongui, and
  Vassilakos]{Catarino2015CervicalChoices}
Rosa Catarino, Patrick Petignat, Gabriel Dongui, and Pierre Vassilakos.
\newblock {Cervical cancer screening in developing countries at a crossroad:
  Emerging technologies and policy choices}.
\newblock \emph{World Journal of Clinical Oncology}, 6\penalty0 (6):\penalty0
  281--290, 2015.
\newblock ISSN 22184333.
\newblock \doi{10.5306/wjco.v6.i6.281}.

\bibitem[Gal and Ghahramani(2016)]{gal2016dropout}
Yarin Gal and Zoubin Ghahramani.
\newblock Dropout as a bayesian approximation: Representing model uncertainty
  in deep learning.
\newblock In \emph{international conference on machine learning}, pages
  1050--1059. PMLR, 2016.

\bibitem[Hinton et~al.(2012)Hinton, Srivastava, Krizhevsky, Sutskever, and
  Salakhutdinov]{hinton2012improving}
Geoffrey~E Hinton, Nitish Srivastava, Alex Krizhevsky, Ilya Sutskever, and
  Ruslan~R Salakhutdinov.
\newblock Improving neural networks by preventing co-adaptation of feature
  detectors.
\newblock \emph{arXiv preprint arXiv:1207.0580}, 2012.

\bibitem[Hu et~al.(2019)Hu, Bell, Antani, Xue, Yu, Horning, Gachuhi, Wilson,
  Jaiswal, Befano, Long, Herrero, Einstein, Burk, Demarco, Gage, Rodriguez,
  Wentzensen, and Schiffman]{Hu2019AnScreening}
Liming Hu, David Bell, Sameer Antani, Zhiyun Xue, Kai Yu, Matthew~P. Horning,
  Noni Gachuhi, Benjamin Wilson, Mayoore~S. Jaiswal, Brian Befano, L.~Rodney
  Long, Rolando Herrero, Mark~H. Einstein, Robert~D. Burk, Maria Demarco,
  Julia~C. Gage, Ana~Cecilia Rodriguez, Nicolas Wentzensen, and Mark Schiffman.
\newblock {An Observational Study of Deep Learning and Automated Evaluation of
  Cervical Images for Cancer Screening}.
\newblock \emph{Journal of the National Cancer Institute}, 111\penalty0
  (9):\penalty0 923--932, 2019.
\newblock ISSN 14602105.
\newblock \doi{10.1093/jnci/djy225}.

\bibitem[IAPB()]{BlindnessVISION2020}
International Agency for the Prevention of~Blindness IAPB.
\newblock {VISION2020}.
\newblock URL \url{https://www.iapb.org:8443}.

\bibitem[Li and Lin(2007)]{li2007ordinal}
Ling Li and Hsuan-Tien Lin.
\newblock Ordinal regression by extended binary classification.
\newblock 2007.

\bibitem[Liberman and Menell(2002)]{Liberman2002BreastBI-RADS}
Laura Liberman and Jennifer~H. Menell.
\newblock {Breast imaging reporting and data system (BI-RADS)}, 2002.
\newblock ISSN 00338389.
\newblock URL \url{https://pubmed.ncbi.nlm.nih.gov/12117184/}.

\bibitem[{MONAI Consortium}(2020)]{the_monai_consortium_2020_4323059}
The {MONAI Consortium}.
\newblock Project monai, December 2020.
\newblock URL \url{https://doi.org/10.5281/zenodo.4323059}.

\bibitem[Paszke et~al.(2017)Paszke, Gross, Chintala, Chanan, Yang, DeVito, Lin,
  Desmaison, Antiga, and Lerer]{paszke2017automatic}
Adam Paszke, Sam Gross, Soumith Chintala, Gregory Chanan, Edward Yang, Zachary
  DeVito, Zeming Lin, Alban Desmaison, Luca Antiga, and Adam Lerer.
\newblock Automatic differentiation in pytorch.
\newblock 2017.

\bibitem[Pisano et~al.(2005)Pisano, Gatsonis, Hendrick, Yaffe, Baum, Acharyya,
  Conant, Fajardo, Bassett, D'Orsi, Jong, and
  Rebner]{Pisano2005DiagnosticScreening}
Etta~D. Pisano, Constantine Gatsonis, Edward Hendrick, Martin Yaffe, Janet~K.
  Baum, Suddhasatta Acharyya, Emily~F. Conant, Laurie~L. Fajardo, Lawrence
  Bassett, Carl D'Orsi, Roberta Jong, and Murray Rebner.
\newblock {Diagnostic Performance of Digital versus Film Mammography for
  Breast-Cancer Screening}.
\newblock \emph{New England Journal of Medicine}, 353\penalty0 (17):\penalty0
  1773--1783, 10 2005.
\newblock ISSN 0028-4793.
\newblock \doi{10.1056/NEJMoa052911}.
\newblock URL \url{http://www.nejm.org/doi/abs/10.1056/NEJMoa052911}.

\bibitem[Ryan et~al.(2014)Ryan, Ostmo, Jonas, Berrocal, Drenser, Horowitz, Lee,
  Simmons, Martinez-Castellanos, Chan, and
  Chiang]{Ryan2014DevelopmentOphthalmology}
Michael~C. Ryan, Susan Ostmo, Karyn Jonas, Audina Berrocal, Kimberly Drenser,
  Jason Horowitz, Thomas~C. Lee, Charles Simmons, Maria~Ana
  Martinez-Castellanos, R.~V.Paul Chan, and Michael~F. Chiang.
\newblock {Development and Evaluation of Reference Standards for Image-based
  Telemedicine Diagnosis and Clinical Research Studies in Ophthalmology}.
\newblock \emph{AMIA ... Annual Symposium proceedings / AMIA Symposium. AMIA
  Symposium}, 2014:\penalty0 1902--1910, 2014.
\newblock ISSN 1942597X.

\bibitem[Schiffman and Solomon(2003)]{schiffman2003findings}
Mark Schiffman and Diane Solomon.
\newblock Findings to date from the ascus-lsil triage study (alts).
\newblock \emph{Archives of pathology \& laboratory medicine}, 127\penalty0
  (8):\penalty0 946--949, 2003.

\bibitem[Srivastava et~al.(2014)Srivastava, Hinton, Krizhevsky, Sutskever, and
  Salakhutdinov]{srivastava2014dropout}
Nitish Srivastava, Geoffrey Hinton, Alex Krizhevsky, Ilya Sutskever, and Ruslan
  Salakhutdinov.
\newblock Dropout: a simple way to prevent neural networks from overfitting.
\newblock \emph{The journal of machine learning research}, 15\penalty0
  (1):\penalty0 1929--1958, 2014.

\bibitem[Xue et~al.(2020)Xue, Novetsky, Einstein, Marcus, Befano, Guo, Demarco,
  Wentzensen, Long, Schiffman, and Antani]{Xue2020ACamera}
Zhiyun Xue, Akiva~P. Novetsky, Mark~H. Einstein, Jenna~Z. Marcus, Brian Befano,
  Peng Guo, Maria Demarco, Nicolas Wentzensen, Leonard~Rodney Long, Mark
  Schiffman, and Sameer Antani.
\newblock {A demonstration of automated visual evaluation of cervical images
  taken with a smartphone camera}.
\newblock \emph{International Journal of Cancer}, 147\penalty0 (9):\penalty0
  2416--2423, 2020.
\newblock ISSN 10970215.
\newblock \doi{10.1002/ijc.33029}.

\end{thebibliography}

\end{document}